\documentclass[letterpaper]{article} % DO NOT CHANGE THIS
\usepackage{aaai25}  % DO NOT CHANGE THIS
\usepackage{times}  % DO NOT CHANGE THIS
\usepackage{helvet}  % DO NOT CHANGE THIS
\usepackage{courier}  % DO NOT CHANGE THIS
\usepackage[hyphens]{url}  % DO NOT CHANGE THIS
\usepackage{graphicx} % DO NOT CHANGE THIS
\usepackage{natbib}  % DO NOT CHANGE THIS AND DO NOT ADD ANY OPTIONS TO IT
\usepackage{caption} % DO NOT CHANGE THIS AND DO NOT ADD ANY OPTIONS TO IT
\usepackage{amsmath} % Added to fix \text in math mode
\usepackage{algorithm}

\usepackage{newfloat}
\usepackage{listings}
\DeclareCaptionStyle{ruled}{labelfont=normalfont,labelsep=colon,strut=off} % DO NOT CHANGE THIS
\floatstyle{ruled}
\newfloat{listing}{tb}{lst}{}
\floatname{listing}{Listing}
\newcommand{\name}{{\sf NoisyHate}}

\title{{\name}:
Mining Online Human-Written Perturbations for Realistic Robustness Benchmarking of Content Moderation Models}

\author {
    Yiran Ye\textsuperscript{\rm 1},
    Thai Le\textsuperscript{\rm 2},
    Dongwon Lee\textsuperscript{\rm 1}
}
\affiliations {
    \textsuperscript{\rm 1}The Pennsylvania State University, USA\\
    \textsuperscript{\rm 2}Indiana University, USA\\
    yiran\_ye@outlook.com, tle@iu.edu, dongwon@psu.edu
}

\usepackage{bibentry}
\usepackage{inconsolata}
\usepackage{graphicx}
\usepackage{booktabs}
\usepackage{makecell}
\usepackage{algorithm}
\usepackage{algpseudocode}
\usepackage{amsfonts}
\usepackage{caption}
\usepackage{subcaption}
\usepackage{soul} % underline
\usepackage{tikz}
\usepackage{multirow}
\usepackage{array}
\usepackage{enumitem}
\usepackage{listings}

\usepackage{xcolor}
\usepackage{url}
\usepackage{float}
\usepackage{listings}

\begin{document}

%New colors defined below
\definecolor{codegreen}{rgb}{0,0.6,0}
\definecolor{codegray}{rgb}{0.5,0.5,0.5}
\definecolor{codepurple}{rgb}{0.58,0,0.82}
\definecolor{backcolour}{rgb}{0.95,0.95,0.92}
\definecolor{cadmiumgreen}{rgb}{0.0, 0.42, 0.24}

%Code listing style named "mystyle"
\lstdefinestyle{mystyle}{
  backgroundcolor=\color{backcolour},
  commentstyle=\color{codegreen},
  keywordstyle=\color{magenta},
  numberstyle=\color{codegray},
  stringstyle=\color{codepurple},
  basicstyle=\ttfamily,
  breakatwhitespace=false,
  breaklines=true,
  captionpos=b,
  keepspaces=true,
  numbers=left,
  numbersep=5pt,
  showspaces=false,
  showstringspaces=false,
  showtabs=false,
  tabsize=2
}

%"mystyle" code listing set
\lstset{style=mystyle}

\maketitle

\begin{abstract}
    Online texts with toxic content are a clear threat to the users on social media in particular and society in general. Although many platforms have adopted various measures (e.g., machine learning-based hate-speech detection systems) to diminish their effect,  toxic content writers have also attempted to evade such measures by using cleverly modified toxic words, so-called {\em human-written text perturbations}. Therefore, to help build automatic detection tools to recognize those perturbations, prior methods have developed sophisticated techniques to generate diverse adversarial samples. However, we note that \ul{these ``algorithm"-generated perturbations do not necessarily capture all the traits of ``human"-written perturbations.} Therefore, in this paper, we introduce a {novel, high-quality dataset of human-written perturbations}, named as {\name}, that was created from real-life perturbations that are both written and verified by human-in-the-loop. We show that perturbations in {\name} have different characteristics than prior algorithm-generated toxic datasets show and thus can be particularly useful to help develop better toxic speech detection solutions. We also provide basic benchmark on the potential utilities of {\name} in perturbation normalization and understanding tasks.
\end{abstract}

\begin{links}
    \link{Code}{https://github.com/YiranYe/NoisyHate}
    \link{Datasets}{https://zenodo.org/records/14654784}
\end{links}

\section{Introduction}\label{sec:intro}
Toxic/hate speech, a conscious and willful defamatory discourse that often targets a specific group of people, has become more prevalent on the Internet. To diminish the online propagation of hate speech, several social media platforms, such as Twitter and Reddit have published content moderation policies specific to hate speech and developed methods to automatically detect and remove them on their platforms. These systems often either utilize a profanity word list and pre-defined rules or train machine learning (ML) algorithms for detecting hate speech. Although these systems have been shown to work well in benchmark test sets, they make one basic assumption that user-generated content online is often written with correct spelling in English. However, we observe that more often trolls do not use English words in their correct forms--e.g., ``bitch", ``stupid'', but misspelled versions of them that can still be understood by humans--e.g., ``bitttch'', ``stupd''. These misspelled words are referred to as text perturbations in ML literature. Because these perturbations are not clean English, they can be used to evade automatic hate speech detection systems online.

\newcolumntype{P}[1]{>{\arraybackslash}p{#1}}
\renewcommand{\tabcolsep}{4pt}
\begin{table}
    \centering
    \footnotesize
    \begin{tabular}{P{0.45\textwidth}}
        \toprule
        \multicolumn{1}{c}{\textbf{Hate Speech Examples from {\name}}} \\
        \cmidrule(lr){1-1}
        % \makecell[l]{
            ``speak for yourself you are an \textbf{\color{red} embarrASSment} (\textbf{\color{cadmiumgreen} embarrassment}) to Canada"
            % }
            \\
            % \makecell[l]{
            \textit{*ASS: Emphasize on ass}
            % }
            \\
        \cmidrule(lr){1-1}
        % \makecell[l]{
            ``right there is a nationwide conspiracy to pick up... arrest \textbf{\color{red} bLAck} (\textbf{\color{cadmiumgreen} black}) men got it"
            % }
            \\
            % \makecell[l]{
            \textit{*LA: a city in the US}
            % }
            \\
        \cmidrule(lr){1-1}
        % \makecell[l]{
            ``then the man is a \textbf{\color{red} p!g} (\textbf{\color{cadmiumgreen} pig}) as well"
            % }
            \\
        \cmidrule(lr){1-1}
        % \makecell[l]{
            ``hey \textbf{\color{red} LOSEEER} (\textbf{\color{cadmiumgreen} loser}) that nut job was a democrat, get over it"
            % }
            \\
        \bottomrule
    \end{tabular}
    \caption{{\name}: Examples of hate speech texts perturbed with \textit{real-life human-written perturbations} (in \textbf{\textcolor{red}{bold, red}}) to benchmark the robustness of content moderation models. The first two texts are distinctive to humans as they contain hidden meanings that can hardly be produced by machines.}
    \label{tab:data_examples}
%    \vspace{-15pt}
\end{table}

These text perturbations have been produced in lots of different ways. One might use \textit{visually similar} characters to replace the original alphabetical characters. For example, ``pr0test" is commonly used to perturb the word ``protest". Another perturbation strategy trolls are likely to use \textit{repeating or removing} certain characters in a word, e.g., ``bitch"$\rightarrow$``bitttch," ``stupid"$\rightarrow$``stupd." Other approaches to produce text perturbations include \textit{placeholder strategy} (``shit"$\rightarrow$``sh\_t"), \textit{lower-upper-case strategy} (``democrats"$\rightarrow$``democRATs"), and the combination of above approaches. These various text perturbations could be a problem that voids the effects of safeguarding ML algorithms.

% Thus trolls sometimes use text perturbations--i.e., misspelled versions of words, to evade such hate speech detection systems.
% Meanwhile, these social platforms also enable the users to block a specific set of words or phrases that they do not want to see, many of which often appear in hate speech or trolling content.
% However, trolls sometimes use another spelling of a potentially offensive word, such as text perturbation, to evade such hate speech detection systems.
% At the same time, a human being can still understand the meaning.
\begin{figure*}
\centerline{\includegraphics[width=1.0\textwidth]{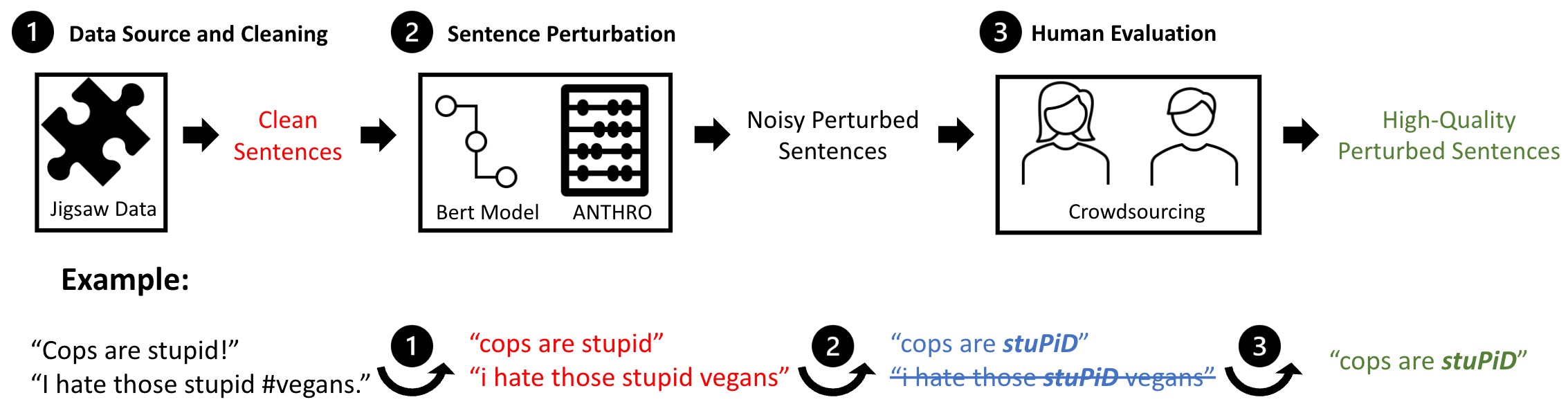}}
\caption{Overall curation pipeline of {\name} dataset. This pipeline has three steps: (1) Data sourcing and cleaning from the original Jigsaw dataset (Section \ref{sec:step1}), (2) Sentence perturbation with human-written perturbations via pseudo-random sampling (Section \ref{sec:step2}) and (3) Human evaluation via crowdsourcing to validate the quality of the perturbed sentences (Section \ref{sec:step3})}
\label{fig:pipeline}
%\vspace{-10pt}
\end{figure*}

Existing works have developed many frameworks to automatically generate text perturbations to benchmark the robustness of ML algorithms~\cite{zeng2020openattack,morris2020textattack}. These frameworks often borrow several adversarial attack algorithms from NLP literature~\cite{yuan2019adversarial,alzantot2018generating,li2018textbugger,zang2020word,li2020bert,gao2018black,ren2019generating,garg2020bae}. However, there is still a gap between those machine-generated perturbations and human-written perturbations~\cite{le2022perturbations}.
% Instead of augmenting datasets with machine-generated perturbations and using them for evaluating the robustness of machine-learning algorithms, we suggest augmenting those datasets with human-written perturbations.
Evaluating ML models using datasets augmented with human-written perturbations is more practical, as it better reflects real-life scenarios. Hence, this paper proposes a benchmark dataset for the toxic speech detection task that contains diverse human-written perturbations online. The contributions of our work are as follows:
\begin{itemize}[leftmargin=\dimexpr\parindent-0.2\labelwidth\relax,noitemsep]
  \item We introduce a novel benchmark test set, {\name}, with online human-written perturbations for toxic speech detection models.
  % This dataset is derived from the popular Jigsaw dataset~\footnote{\url{https://jigsaw.google.com/}} with toxicity labels and identity annotations.
  \item Our evaluation with state-of-the-art (SOTA) language models and the commercial toxic detection \textit{Perspective API}~\footnote{\url{https://perspectiveapi.com/}} on {\name} reveals that there is still room for improving the robustness of these models on predicting texts with human-written perturbations.
  \item We test {\name} dataset with several spell checkers and show that it is worth developing a better normalization tool targeting these online human-written perturbations.
\end{itemize}

\section{Related Work}

 % Many researchers have made significant contributions to the toxic speech detection area, as evidenced by the proliferation of well-structured and widely utilized datasets such as \textit{HatEval} 2019~\cite{basile2019HatEval}, \textit{OffensEval} 2019~\cite{zampieri2019OffensEval} and 2020\cite{zampieri2020OffensEval}, and \textit{TweetEval} 2020 \cite{barbieri2020tweeteval}. However, these datasets often contain only clean English texts and might not reflect the reality that many toxic texts are written not in clean, correctly spelled but perturbed forms. Evidently, \textsc{CrypText}~\cite{le2023cryptext} collects human-written perturbations directly from social media sites such as Reddit and shows that there are many text perturbations written by humans that closely resemble correct-spelled toxic words in terms of how they sound, spell and also meanings. These perturbations are very different from ones generated by computer algorithms in existing adversarial text literature. Most of the existing algorithms either modify the spellings of words\cite{bhalerao2022data, li2018textbugger, gao2018black, eger2019text, hosseini2017deceiving} or substitute words with their synonyms \cite{alzantot2018generating, ribeiro2018semantically, sato2018interpretable, jin2020bert}. However, these text perturbation strategies are often deductively derived~\cite{le2022perturbations} based on some known vulnerabilities of ML classification models. Thus, how well they can help the hate-speech detection model prevent real-world attacks are yet to be explored.

Two research areas that are closely related to our work are text perturbation generation algorithms and toxic speech detection.

\subsection{Text Perturbation Generation}
\noindent {\em Machine Generated Text Perturbations.} In literature, two major approaches are used to generate adversarial text samples: spelling modification and close word substitution \cite{alzantot2018generating, ribeiro2018semantically, sato2018interpretable, jin2020bert}. The spelling modification approach usually involves deleting, inserting, swapping, and replacing certain characters in a word. Bhalerao et al. \cite{bhalerao2022data} proposed a tool named Continuous Word2Vec (CW2V) to perturb text with the following rules: Fake punctuation (``like"$\rightarrow$``l.i.k.e"), Neighboring key (``like"$\rightarrow$``lime"), Random spaces (``like"$\rightarrow$``l ike"), Transposition (``share"$\rightarrow$``sahre"), and Vowel repetition and deletion (``like"$\rightarrow$``likee"). Other than changing the word's spelling directly, the second approach aims to replace the entire word with other regular English words to attack text classification models. Ribeiro et al.'s work \cite{ribeiro2018semantically} demonstrated the effectiveness of the attack with semantically equivalent adversarial rules (SEARs) on machine comprehension, visual question answering, and sentiment analysis tasks. SEARs are simple universal replacement rules intending to convert the target word into its semantically identical pairs (``what"$\rightarrow$``which", ``what is"$\rightarrow$``what's"). \citealp{alzantot2018generating} also introduced an adversarial attack method that replaces the target word with its top k nearest neighbors based on the distance in the GloVe embedding space. To make the perturbation types more diverse, \citealp{li2018textbugger} developed TEXTBUGGER, which applied both the spelling modification and close word substitution approaches.
Nevertheless, experiments \cite{le2022perturbations} involving human evaluation reveal that the distance between the perturbations generated by machines and real Human-written Text Perturbations still exists~\cite{le2022perturbations}. Moreover, since these adversarial samples are produced based on some known vulnerabilities of a target model, how well they can help the hate-speech detection model prevent real-world attacks is yet to be explored.

\vspace{2pt}
\noindent {\em Human-written Text Perturbations.} \textsc{CrypText}~\cite{le2023cryptext} is a platform that retrieves human-written perturbations directly from social media, such as Reddit, and provides visualization on the trend of those perturbations. Meanwhile, it also offers an interface to perturb the user-inputted sentences randomly.
Due to this randomness, some insignificant words might be perturbed occasionally. For example, given a sentence, ``I hate those stupid vegans", a perturbation on ``I" or ``those" might be insufficient to help this sentence evade the hate speech detection system.
Moreover, the \textsc{CrypText} is using the \textsc{Anthro} algorithm~\cite{le2022perturbations} to cluster the words and their perturbations in social media based on their sound and spelling composition (e.g., leading characters, vowels and consonants, and visually similar characters). Therefore, some different standard English words with the same pronunciation, such as ``maim" and ``mam," will be treated as each other's perturbation. Hence, the randomly picked perturbation might not fit the context well.

\vspace{2pt}
\noindent \textbf{Toxic Speech Detection Dataset.} Many researchers have made significant contributions to the toxic speech detection area, as evidenced by the proliferation of well-structured and widely utilized datasets, HatEval 2019\cite{basile2019HatEval}, OffensEval 2019 \cite{zampieri2019OffensEval} and 2020\cite{zampieri2020OffensEval}, and TweetEval 2020 \cite{barbieri2020tweeteval}. However, these datasets often contain only clean English texts.
throughout this period. Barbieri et al. \cite{barbieri2020tweeteval} tested pre-trained Transformers, such as BERT, RoBERTa, and other popular models, such as LSTM and SVM, on the TweetEval data set and reported that the best Macro F1 score on the test set is 0.829 (RoBERTa-Retrained) for offensive speech classification. Mathew et al. \
~\cite{mathew2021hatexplain} introduced another dataset named HateXplain and also trained language models, including BERT and BiRNN, on this dataset.
According to their report, BERT demonstrated the best Macro F1 result, 0.687.
Perspective API \cite{perspectiveapi}, created by the Jigsaw and Google team, is one of the most famous black box toxic content detection systems.
According to their website, their model was trained on millions of comments from various sources, including online forums such as Wikipedia and The New York Times, across various languages.
It is still worthwhile to investigate the performance of these models when subjected to human-written perturbations.

\begin{listing}[tb]%
\caption{Python code for loading the {\name} datasets into a table using the Hugging Face API}%
\label{lst:python_code}%
\begin{lstlisting}[language=Python]
from datasets import load_dataset

url = "NoisyHate/Noisy_Hate_Data"
clean_set = load_dataset(url,
  data_files={"clean":"data/clean.csv"})
pert_set = load_dataset(url,
  data_files={"pert":"data/pert.csv"})
\end{lstlisting}
\end{listing}

\section{{\name} Dataset}

\subsection{Overview and Usage}
This section introduces the transformation pipeline from the Jigsaw dataset's original texts to {\name} of three consecutive steps, namely (1) data pre-processing, (2) sentence perturbation using human-written perturbations, and (3) quality assurance via crowdsourcing (Fig. \ref{fig:pipeline}). {\name} dataset is also accessible through HuggingFace's data repository~\footnote{\url{https://huggingface.co/datasets/NoisyHate/Noisy_Hate_Data}} or the public DOI: \textit{10.5281/zenodo.14654784}. Listing \ref{lst:python_code} shares the code snippets to retrieve and load the {\name} into a table format using Python. This semi-automatic pipeline hopes to provide researchers with a consistent framework, including user-study designs and interfaces to curate benchmark datasets with human-written perturbations in domains other than toxic text detection.

% \begin{figure}[t]
%   \centering
%   \includegraphics[width=\linewidth]{emnlp2023-latex/figs/anonymous_code_short.jpg}
%   \caption{Python code for loading the {\name} datasets into a table using the Hugging Face API}
% \label{fig:python_code}
% % \vspace{-5pt}
% \end{figure}

\subsection{Step 1: Data Source and Cleaning}\label{sec:step1}

Jigsaw is a popular toxic speech detection dataset of nearly 2M public comments from the Civil Comments platform. In addition to the toxic score labels for toxicity classification, it also provides several toxicity sub-type that indicate particular comments' target groups. Due to these prolific identity annotations and significant data volume, we adopted this dataset as our raw data source. There also exist many other datasets, such as HateXplain \cite{mathew2021hatexplain} and Measuring Hate Speech \cite{social_sciences_data_lab_at_uc_berkeley_2024}. However, they are generated and annotated based on different methodologies and instructions; thus, incorporating these datasets might introduce inconsistencies or biases to our evaluation. Hence, we used only Jigsaw. Moreover, since the dataset has been used as the standard benchmark dataset for several content moderation tasks, this adoption will also help reduce the entry barrier in {\name}'s adoption. Since the comments from the Jigsaw dataset contain a lot of special characters, emojis, and informal language, data cleaning was necessary to ensure data quality. Following a typical text processing pipeline, we removed duplicated texts, special characters, special punctuation, hyperlinks, and numbers. Since we only focused on \textit{positive (hate speech)} English texts, sentences containing non-standard English words were filtered out. 13,1982 comments remained after this cleaning step.

\begin{algorithm}[t]
        \small
	\caption{Perturbing Process}
	\begin{algorithmic}[1]
	\Require Input $S_{clean}{=}\{w_{1},w_{2},...,w_{n}\}$,
     BERT model $B$,
     Perturbations Generated by the ANTHRO Algorithm $Dict:(w,[p_{1},p_{2},...,p_{t}])$,
     Output Size $k$, Threshold $\Theta$
	    \State $\mathbb{S}_{pert}=[]$;\;\;$Score_{origin}=B(S_{clean})$
		% \State
  %\Comment Calculate the original toxic score
		\For {$w_{i}$ in $S_{clean}$}
    		\State $Score_{mask_{i}}=B([w_{1},w_{2},...,w_{i-1},w_{i+1},...,w_{n}])$
      %\Comment Calculate the toxic score with mask on $w_{i}$
    		\If {$abs(Score_{origin}-Score_{mask_{i}})>\Theta$}
        		\For {$p_{j}$ in $random.sample(Dict_{hard}[w_{i}],k)$}
        		    \State $\mathbb{S}_{pert}$ += $[w_{1},w_{2},...,w_{i-1},p_{j},w_{i+1},...,w_{n}]$
        		\EndFor
    		\EndIf

		\EndFor
        \State \textbf{Return:} Perturbed Sentence list $\mathbb{S}_{pert}=[S_{1},S_{2},...,S_{k}]$
	\end{algorithmic}
	% he is a fucking gay -> [he is a fucking, he is a gay]
	%sampling a subset such that one perturbation do not appear too many times -> decrease theta
	%sampling with probability as the abs in score difference -> so if the difference in score is high => more chance to be selected
	%np.random.sample(list, p=XXX)
	\label{alg:perturb}
\end{algorithm}

\subsection{Step 2: Sentence Perturbation}\label{sec:step2}

In this step, we aim to perturb each of the comments resulting from Step 1 with human-written perturbations. The pseudo-code of this sentence perturbation process is described in Alg. \ref{alg:perturb}.

\vspace{2pt}
\noindent \textbf{Word Importance Scoring.} Since we focus on benchmarking toxic text detection tasks, it is practical and meaningful to perturb only a few critical words within a sentence. To do this, we first train a proxy toxic detector and utilize it to approximate the importance of each word to toxicity prediction. In particular, we first fine-tune a BERT model~\cite{devlin2018bert} on the Jigsaw dataset. Then, we enumerate and mask every words in each sentence and observe how much confidence the trained model changes after such masking operations (Alg. \ref{alg:perturb}, Line 4--Line 9). A candidate word is selected to be perturbed at every enumeration step if masking decreases the proxy model's confidence more than a pre-determined threshold. (Alg. \ref{alg:perturb}, Line 5). In this experiment, 0.7 was selected to be the pre-determined threshold.

\vspace{2pt}
\noindent \textbf{Word Selection and Perturbation.} A comment might have more than one crucial word. However, perturbing all of them will increase the chance that the perturbed sentence being discarded in the annotators.
% This can happen because one poorly perturbed word can lead to the discard of the whole sentence.
Thus, to maximize the number of comments that remained at the end, we take a conservation measure and only perturb the most important word in a given sentence. Then, we utilize the \textsc{Anthro} algorithm~\cite{le2022perturbations}, which provides the cluster of online human-written perturbations for a given word, to retrieve the perturbations from social media on the selected important words.

\renewcommand{\tabcolsep}{1pt}
\begin{table}
    \centering
    \footnotesize
    \begin{tabular}{lcr}
        \toprule
        \textbf{Type} & \textbf{Description} & \textbf{Example} \\
        \cmidrule(lr){1-3}
        RepeatChar & \makecell[l]{
            repeat several characters
            } & stupid{$\rightarrow$}stuppppid \\
        \cmidrule(lr){1-3}
        Abbr & \makecell[l]{
            delete several characters
            } & stupid{$\rightarrow$}stupd\\
        \cmidrule(lr){1-3}
        SpecialChar & \makecell[l]{
            replace several characters \\with Non-English one\\
            } & \makecell[r]{stupid{$\rightarrow$}5tupid \\ stupid{$\rightarrow$}st*pid}\\
        \cmidrule(lr){1-3}
        MixedCase & \makecell[l]{
            replace several characters \\with their upper cases\\
            } & stupid{$\rightarrow$}sTuPId\\
        \cmidrule(lr){1-3}
        MixedCase+ & \makecell[l]{
            replace several characters\\ with their
            upper cases, while\\ these characters can
            be \\combined into a new word
            } & stupid{$\rightarrow$}stuPiD\\
        \bottomrule
    \end{tabular}
    \caption{We categorize all human-written perturbations into five different groups according to five perturbation strategies that curate them. Noticeably, we may encounter more complex perturbations, such as hybrid types (e.g., ``shit"$\rightarrow$``sh!!!!!t"). In this case, we will mark ``sh!!!!!t" as both RepeatChar and SpecialChar.}
    \label{tab:types}
    % \vspace{-25pt}
\end{table}

\vspace{2pt}
\noindent \textbf{Perturbation Type Diversification.} Different types of human-written perturbations exist according to the perturbation strategies that curate them. To categorize all the perturbations used in this step, we classify them into five different perturbation strategies. Table \ref{tab:types} presents the definition of those types and examples. They are (1) repeating characters (RepeatChar), (2) using abbreviation by removing one or several characters (Abbr), (3) using non-English or special characters (SpecialChar), (4) using mixed cased characters (MixedCase) and (5) using mixed cased characters with an additional layer of meaning (e.g., ``republicans"$\rightarrow$``repubLIEcans") (MixedCase+). Due to the imbalanced frequency distribution among different perturbation strategies, the \textsc{Anthro} algorithm~\cite{le2022perturbations} is biased to a specific type of perturbation, such as RepeatChar. Hence, to increase the diversity of different perturbation types in {\name} dataset, we apply a pseudo-random strategy to give a higher chance for perturbation types with less frequency to be sampled. We utilized this procedure with ten random seeds and chose the best distribution entropy. This step resulted in 2,120 comments. Even though this reduced the total number of comments to a great extent, we prioritized diversity and balance across perturbation types, which are crucial for robust evaluation. The resulting subset offers a more representative and challenging benchmark for testing model resilience against various perturbations.

% \begin{figure}
% \centerline{\includegraphics[width=\linewidth]{figs/Perturbation Type Frequency Pie Chart Before.png}}
% \caption{Perturbation Type Frequency Pie Chart Before Human Evaluation}
% \label{fig:type_dist_before}
% \vspace{-5pt}
% \end{figure}

\subsection{Step 3: Quality Assurance via Crowd-sourcing}\label{sec:step3}

\begin{figure*}[h]
\centerline{\includegraphics[width=0.9\linewidth]{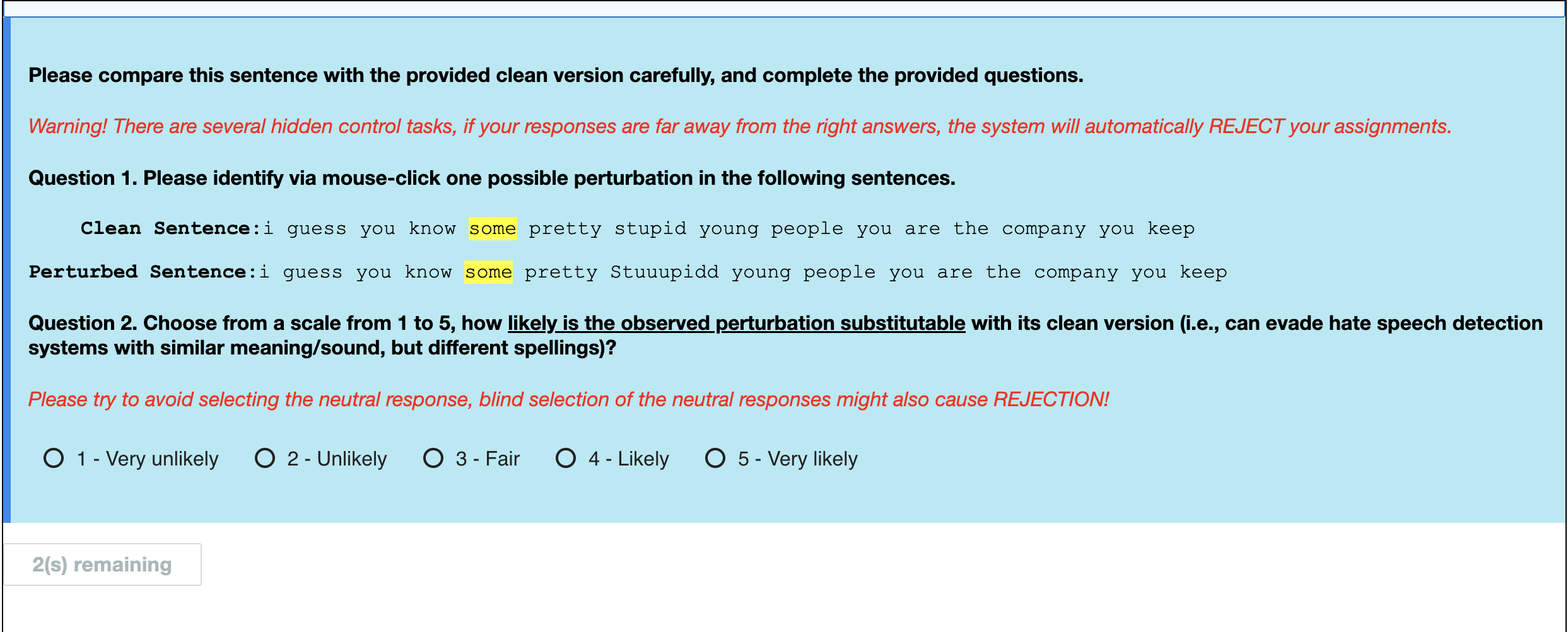}}
\caption{Human evaluation Interface: a clean-perturbed word pair will be highlighted when the worker moves the mouse cursor over one of them. By clicking the highlighted word, the worker commits that this is the identified clean-perturbed pair.}
\label{fig:interface}
\end{figure*}

\vspace{2pt}
\noindent \textbf{Training-Based Crowdsourcing.} With the dataset of perturbed comments prepared in Section \ref{sec:step2}, we proceed to the human evaluation step that involves Amazon Mechanical Turk (MTurk) workers in judging the generated perturbation's quality. This step is to ensure that the select perturbations in the previous step (Section \ref{sec:step2}) are appropriate to the sentences' contexts. Before MTurk workers work on their tasks, we provide brief training to the workers by providing them with a few examples. To do this, we display a guideline to explain the definition of human-generated perturbation and provide examples of both high-quality and low-quality perturbations (Appendix, Table \ref{tab:instruction}). This training phase has been suggested to warrant high-quality responses from the human worker, especially for labeling tasks~\cite{clark2021all}. Each MTurk worker is then presented with a pair of perturbed sentences, its clean version, and asked to determine the quality of the perturbed one (Appendix, Fig. \ref{fig:interface}).

\vspace{2pt}
\noindent \textbf{Recruitment and Compensation.} We recruited five different workers from the North American region through five batches to assess each pair. A five-second countdown timer was also set for each task to ensure that workers spent enough time on it. To ensure the quality of their responses, we designed an attention question that asks them to click on the perturbed word in the given sentences before they provide their quality ratings (Appendix, Fig. \ref{fig:interface}). Workers who cannot correctly identify the perturbation's location in the given sentence will be blocked for future batches. We aimed to pay the workers at an average rate of \$10 per hour, which is well above the federal minimum wage of \$7.25 per hour. The payment of each task was estimated by the average length of the sentences, which totals around 25 words per pair, and the average reading speed of native speakers is around 228 words per minute~\cite{trauzettel2012standardized}. Our study was approved by an institutional IRB board.
% Hence, for each task, we paid $ \$10/(60*228/25) \approx \$0.02 $.

\begin{figure}[tb]
     \centering
    \small
     \begin{subfigure}[b]{0.45\textwidth}
         \centering
         \includegraphics[width=\textwidth]{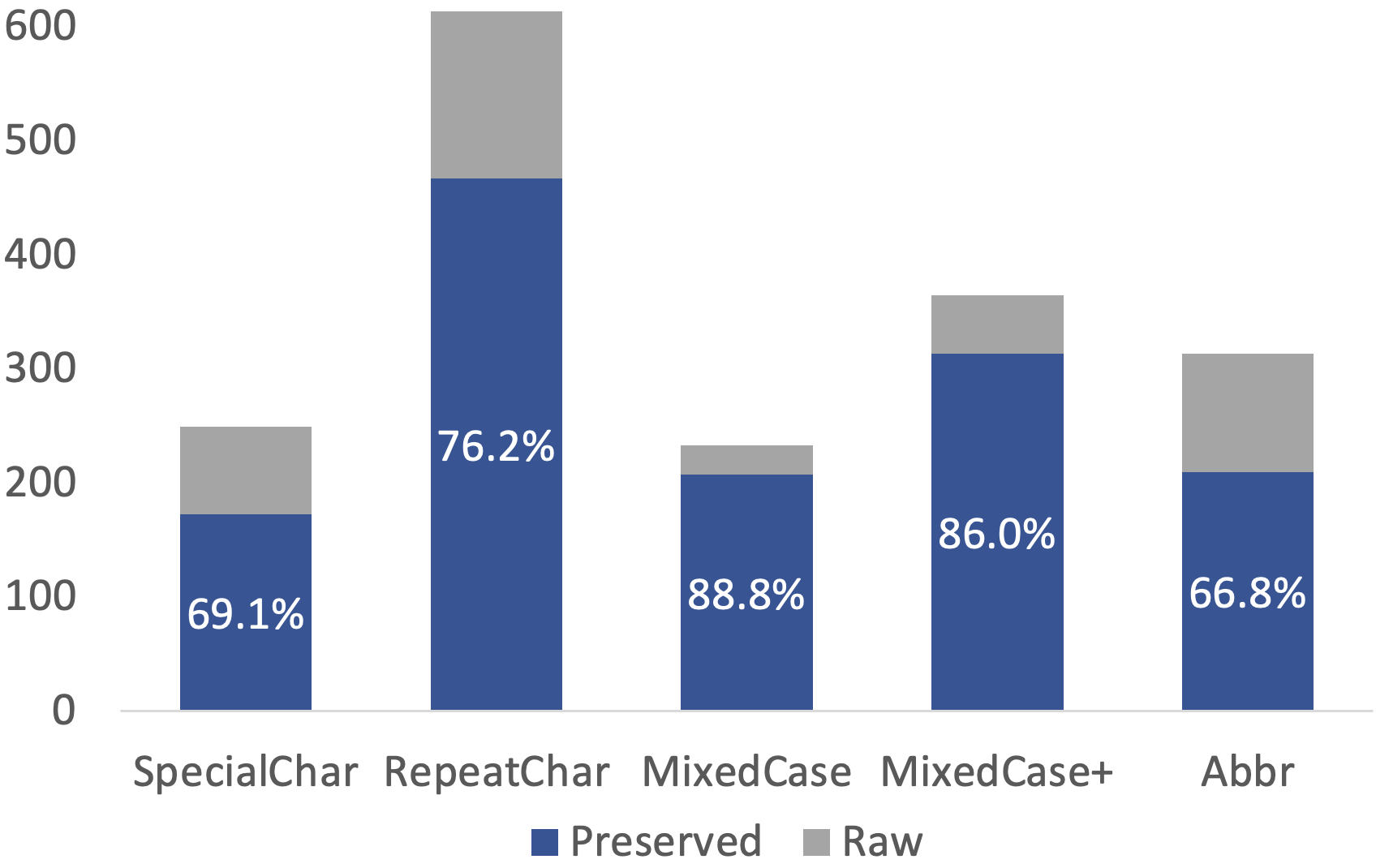}
         \caption{Percentage of perturbation remained across all five perturbation types.}
         \label{fig:a}
     \end{subfigure}
     \hfill
     \begin{subfigure}[b]{0.45\textwidth}
         \centering
         \includegraphics[width=\textwidth]{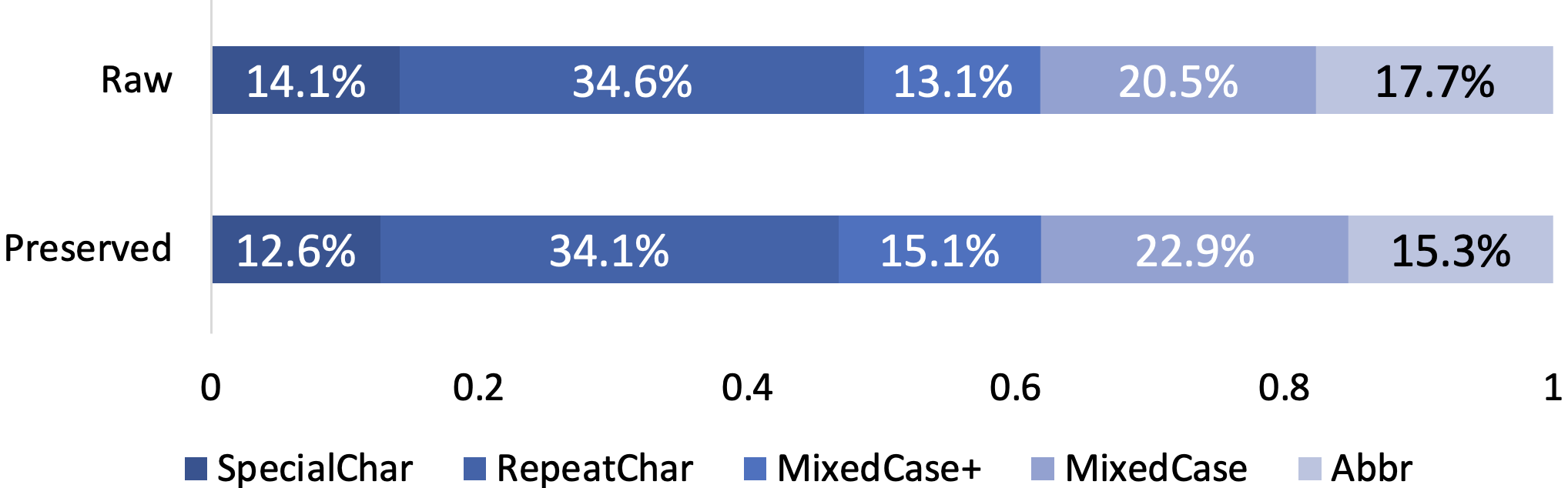}
         \caption{Perturbation type distribution}
         \label{fig:b}
     \end{subfigure}
        \caption{Comparison of the distribution of different perturbation categories \textit{before and after} validated by human workers (Section \ref{sec:step3}). ``Raw" refers to the original data we send to MTurk workers, and ``preserved" refers to the number/percentage saved after human evaluation.}
        \label{fig:type_dist}
        % \vspace{-30pt}
\end{figure}

\vspace{2pt}
\noindent \textbf{Quality Assurance.} By removing tasks that failed to identify the location of the perturbed words accurately, the number of comments in the dataset was reduced from 2,120 to 1,707 examples. Subsequently, approximately 78.4\% (1,339 examples) of the remaining data were deemed high-quality perturbations and retained for further analysis. Even though 22\% of the data were removed by crowdsourcing workers and the number seems to be small, it still meets our expectations as all the perturbations were collected from the internet and our data should make more sense to humans. Although 1,339 appears to be a small size for a dataset, we intend {\name} to be a public benchmark test set. Hence, if we use the general train-test split ratio, 9:1, a size of 1,339 is comparable to the size of the test set of current popular data sets, such as OffensEval-2019 (13,240 examples) and SemEval-2021 (10,000 examples). Fig. \ref{fig:type_dist} presents the distribution of high-quality perturbations across five categories. MixedCase+ category (e.g., ``republicans"{$\rightarrow$}``repubLIEcans") had the highest retention rate of 88.8\%.
% Conversely, MTurk workers discarded more than 30\% of perturbed sentences containing abbreviation perturbations or special characters—the outcome of our study aligned with our initial expectations.
The generation of MixedCase+ perturbations typically requires human insights and understanding of outside contexts that are not accessible by machines, which makes them more acceptable to humans. In contrast, both abbreviation-based perturbations (Abbr) and those utilizing special characters (SpecialChar) result in character loss, making it more challenging for humans to associate the perturbed sentence with its clean version, resulting in a 30\% discard rate.

% \renewcommand{\tabcolsep}{5pt}
% \begin{table*}[tb]
%     \centering
%     \footnotesize
%     \begin{tabular}{lr}
%         \toprule
%         \textbf{Perturbation Type} & \textbf{Description} & \textbf{Example} \\
%         \cmidrule(lr){1-2}
%         repeat\_char & \makecell[l]{
%             Perturbation that is generated by repeating several characters\\ in its original clean version.
%             } & Stupid -> Stuppppid \\
%         \cmidrule(lr){1-2}
%         abbr & Perturbation that is generated by deleting several characters in its original clean version. & Stupid -> Stupd\\
%         \cmidrule(lr){1-2}
%         special\_char & Perturbation that is generated by deleting several characters in its original clean version. & Stupid -> Stupd \\
%         \cmidrule(lr){1-2}
%         lower\_upper\_case & Perturbation that is generated by deleting several characters in its original clean version. & Stupid -> Stupd \\
%         \cmidrule(lr){1-2}
%         interesting\_lower\_upper\_case & Perturbation that is generated by deleting several characters in its original clean version. & Stupid -> Stupd \\
%         \bottomrule
%     \end{tabular}
%     \caption{Perturbation Types}
%     \label{tab:example}
% \end{table*}

\section{Potential Uses}
This section evaluates {\name}'s dataset on two NLP tasks that correspond to research questions (RQs) as follows.
% Answers to this question can demonstrate the novelty of the proposal of using real-life human-written perturbations

\vspace{2pt}
\noindent \textbf{RQ. 1. (Perturbation Normalization)} \textit{How much can {\name}'s perturbed sentences be restored to their clean version by existing normalization algorithms such as misspelling correctors?} \\
% Answers to this question can demonstrate the novelty of real-life human-written perturbations.\\
\textbf{RQ. 2. (Perturbation Understanding)} \textit{How effectively can {\name}'s perturbed sentences attack state-of-the-art deep-learning-based and commercial toxic detection models?}
% Answers to this question can demonstrate the potential impact of {\name} dataset on improving future toxic text detection models.

\subsection{Perturbations Normalization}
\noindent \textbf{Experiment Set-up.} To answer RQ. 1, we select one word-level spell corrector: pyspellchecker \cite{pyspellchecker}, one open source deep-learning-based misspell corrector: NeuSpell \cite{jayanthi-etal-2020-neuspell}, and two commercial APIs: Google Search SerpApi~\cite{googlespell} and Bing Spell Checker API~\cite{bingspell}. We detailed these spelling correctors below.

\begin{figure*}[hbt!]
    \centering
    \includegraphics[width=0.98\linewidth]{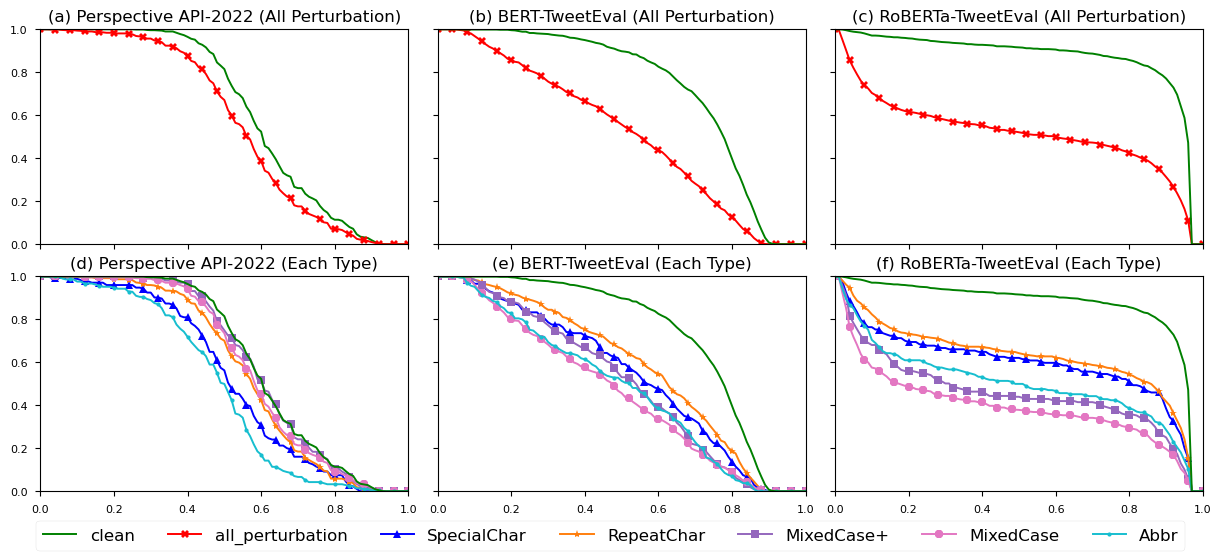}
    % \vspace{-10pt}
    \caption{Model accuracy vs. threshold curves: in each chart, the x-axis represents the threshold, and the y-axis is the models' accuracy. In the top row charts, the all\_perturbation curve represents the weighted mean accuracy, computed by combining the accuracy of each perturbation type weighted by its frequency.}
    \label{fig:understanding}
    % \vspace{-5pt}
\end{figure*}

\begin{itemize}[leftmargin=\dimexpr\parindent-0.2\labelwidth\relax]
  \item \textit{pyspellchecker.} pyspellchecker is a word-level misspelling checker that returns the target word $w$'s most likely permutations within a predefined Levenshtein Distance.
  % This likelihood for a candidate permutation $c$ is calculated based on the multiplication of its word frequency, $P(c)$, and the probability that the $w$ is typed when the author meant to type the candidate $c$, $P(w|c)$.
  \item \textit{NeuSpell.} NeuSpell is an open-source toolkit for sentence-level spelling correction. It contains 10 language models, including CNN-LSTM, Nested-LSTM, and BERT. We test all models on our dataset and report the best performance.
  \item \textit{Google Search SerpApi} Google Search provides ``Did you mean?" suggestions when a user inputs a sentence with spelling errors in its search bar. SerpApi extracts these suggestions and provides them as a spell-checker API.
  \item \textit{Bing Spell Checker API.} The Bing Spell Checker API utilizes statistical machine translation and ML algorithms to deliver contextual corrections. According to its website, it can recognize and normalize slang, informal language, and different words with the same sounds (``see" and ``sea").
\end{itemize}

\renewcommand{\tabcolsep}{0.1pt}
\begin{table}[tb!]
    \centering
    \small
    \begin{tabular}{lcccccc}
        \toprule
        \textbf{Spell} & \textbf{Repeat-} & \textbf{Abbr} & \textbf{Special-} & \textbf{Mixed-} & \textbf{Mixed-} & \textbf{Overall} \\
        \textbf{Corrector} & \textbf{Char} & & \textbf{Char} & \textbf{Case} & \textbf{Case+} & \\
        \cmidrule(lr){1-7}
        Google & \textbf{0.698} & \textbf{0.751} & 0.322 & 0.934 & 0.954 & \textbf{0.755}\\
        Bing API       & 0.406 & 0.533 & 0.426 & 0.943 & \textbf{0.963} & 0.633\\
        pyspellchecker & 0.544 & 0.480 & \textbf{0.809} & \textbf{0.948} & 0.954 & 0.728 \\
        NeuSpell       & 0.294 & 0.550 & 0.448 & 0.867 & 0.931 & 0.583 \\
        \bottomrule
        % \multicolumn{6}{l}{(*) via SerpApi}
    \end{tabular}
    \caption{Perturbation Normalization Accuracy}
    \label{tab:spellcorrector}
    % \vspace{-20pt}
\end{table}

\noindent \textbf{Experiment Results.}
% {\name}'s perturbed sentences contain lowercase and uppercase characters. However, our clean set only has lowercase characters; some spell correctors might choose to retain the letter case (e.g., ``Stupiid" will be corrected as ``Stupid" instead of ``stupid"), a corrected word with uppercase characters might be treated as failure even if it has the same spelling as the clean word. In this case, we convert the corrected word into lowercase before comparison. Noticeably, this operation might reduce the toxicity of certain words (``democRATs"{$\rightarrow$}``democrats").
Table \ref{tab:spellcorrector} summarizes the accuracy of these spell checkers on {\name}. Although two of the five perturbing methods (MixedCase and MixedCase+) allow normalization accuracy close to or over 95\%, at least two reasons suggest we retain those perturbations. First of all, blindly applying such normalization is unrealistic since this operation will reduce the toxicity score and even change toxicity polarity. In addition, BERT-based models are sensitive toward letter cases (we further discussed this in Section \ref{hateDetectionResults}). By retaining MixedCase and MixedCase+, we are expecting novel hate speech detection models that can perform normalization as well as preserve the potential meaning of the perturbed word carrying (e.g., ``rats" at ``democRATs"). Overall, one of the commercial APIs, Google Search, presents the best result. Surprisingly, the word-level spell corrector, pyspellchecker, has the second-highest accuracy. One possible explanation for this observation is that, for an inputted non-English word, the pyspellchecker always offered its best guess even if the calculated probability of typos was low, while other tools remained conservative with a small confidence value.
% In practice, one sentence could contain more than one perturbation.
In practice, these spell checkers might encounter more complex and novel perturbations. This result calls for continuous improvement of normalization tools for human-written perturbations online.

\renewcommand{\tabcolsep}{1pt}
\begin{table}[tb]
    \centering
    \small
    \footnotesize
    \begin{tabular}{llcccccc}
        \toprule
        \textbf{Model} & \textbf{} & \textbf{Repeat} & \textbf{Abbr} & \textbf{Special} & \textbf{Mixed} & \textbf{Mixed} & \textbf{AllPert} \\
        \textbf{} & {} & \textbf{Char} & & \textbf{Char} & \textbf{Case} & \textbf{Case+}\\
        \cmidrule(llrrrrrr){1-8}
        \multirow{2}{4.2em}{Perspective API} & $\Delta_{\text{AUC}}$ & 0.047 & 0.138 & 0.092 & 0.025 & 0.009 & 0.057\\
        & $\Delta_{ACC_{0.5}}$ & -0.113 & -0.314 & -0.251 & -0.078 & -0.065 & -0.142\\
        \cmidrule(llrrrrrr){1-8}
        \multirow{2}{4.2em}{BERT-TweetEval} & $\Delta_{\text{AUC}}$       & 0.149 & 0.248 & 0.188 & 0.267 & 0.222 & 0.151\\
        & $\Delta_{ACC_{0.5}}$ & -0.232 & -0.383 & -0.305 & -0.44 & -0.366 & -0.339\\
        \cmidrule(llrrrrrr){1-8}
        \multirow{2}{4.2em}{RoBERTa-TweetEval} & $\Delta_{\text{AUC}}$    & 0.244 & 0.369 & 0.275 & 0.474 & 0.414 & 0.212 \\
        & $\Delta_{ACC_{0.5}}$ & -0.272 & -0.415 & -0.294 & -0.536 & -0.475 & -0.392\\
        \bottomrule
    \end{tabular}
    \caption{Classification results of three hate speech detectors for each perturbation type. $\Delta_{AUC}$ is the gap area between the perturbed set and the clean set. $ACC_{0.5}$ is the classification accuracy at $t=0.5$, and $\Delta_{ACC_{0.5}}$ refers to the degradation from clean accuracy by certain attacks.}
    \label{tab:auc_diff}
    % \vspace{-25pt}
\end{table}

\subsection{Perturbations Understanding} \label{hateDetectionResults}

\noindent \textbf{Experiment Set-up.} For \textbf{RQ. 2.}, we tested the BERT model~\cite{devlin2018bert}, RoBERTa model~\cite{liu2019roberta} and also the Perspective API on {\name} dataset. Perspective API \cite{perspectiveapi}, created by the Jigsaw and Google team, is one of the most popular toxic content detection APIs.
% Researchers also studied algorithmic adversarial attacks targeting Perspective API (e.g., \cite{jain2018adversarial, hosseini2017deceiving, gil2019white, brown2019acoustic}). However, as its version iterates, Perspective API developed defensive strategies to thwart these machine-generated attacks.
% Table \ref{tab:perspectiveattackfull} presents the outcomes of today's Perspective API when facing previously effective attacks generated using machines.

We compare the performance of the detectors when evaluated on both clean and perturbed sentences. Since all of the data in our {\name} dataset are positive examples (i.e., toxicity$\geq$0.5 in the original Jigsaw dataset), the model's classification accuracy depends on a pre-defined decision threshold $t$. Typically, $t{=}k$ means that the input will be considered toxic if the probability of a model's prediction for this data is larger than $k$, and vice versa. To better capture the performance differences among different threshold values, we plot the model accuracy vs. threshold curve as shown in Fig. \ref{fig:understanding}. Meanwhile, two measures, $\Delta_{AUC}$ and $ACC_{0.5}$ were applied to evaluate the model's performance. $\Delta_{AUC}$ defines the gap between the area under the perturbed curve and the clean curve.
% , calculated as Equation \ref{eq:auc} where $C(t)$ refers to the classification accuracy on clean set at threshold $t$ and $P(t)$ refers to the classification accuracy on perturbed set at threshold $t$,
The lower the $\Delta_{AUC}$ value is, the more robust a model is. $ACC_{0.5}$ refers to the model's classification accuracy when the threshold $t$ is equal to 0.5.

% \begin{equation}
% \Delta_{AUC} = \int_{0}^{1} (C(t) - P(t)) dt
% \label{eq:auc}
% \end{equation}

\vspace{2pt}
\noindent \textbf{Experiment Results.} Table \ref{tab:auc_diff} summarizes the results. We observe that the RoBERTa-TweetEval model has the best performance ($ACC_{0.5}{=}0.915$) on the clean set while the Perspective API has the worst performance ($ACC_{0.5}{=}0.814$). However, the perspective API achieves the best overall robustness compared to other models. In the perturbed dataset, the accuracy of the Perspective API's classification when $t{=}0.5$ is 0.672, which is much higher than the second best model, the BERT model ($ACC_{0.5}{=}0.557$). In addition, the Perspective API attains the smallest gap area, $\Delta_{AUC}{=}0.057$, between the perturbed set and the clean set (Table \ref{tab:auc_diff}). This also indicates that the Perspective API is least affected by perturbations. Although the Perspective API and the Jigsaw dataset have been developed by the same research group, model overfitting is still the least of the concerns since it has the worst performance on the clean dataset. We tend to believe it's a trade-off the Perspective API, as a commercial API, made between robustness and accuracy.

\section{Limitations}
Our work has two main limitations. First, the toxicity of a sentence can evolve over time. For instance, words that may have been considered neutral several decades ago, such as ``retarded'' may be perceived as toxic in contemporary times. Also, people might generate more interesting and novel perturbations in the future and, hence will require updating our dataset continually.
% As for future work, we aim to develop a model that can distinguish high-quality perturbations based on the annotated data we collected from Amazon MTurk. This model will enable us to automate the perturbation process, ensuring that the perturbations remain timely and relevant.
Second, within the limit of a conference paper, we strictly focus on introducing two potential uses: perturbation normalization and understanding. However, {\name} can also be used as additional training examples to improve existing toxic detection models via methods such as adversarial training. We foresee no potential risks of our dataset to the community and society.
% our decision of not including such analysis does not lessen \textsc{NoisyHate}'s value as a novel and realistic toxic benchmarking dataset that is curated with real human-written perturbations online, which is the main contribution of our paper.

\section{Conclusions and Future Work}

This work presents a novel data pipeline with a human-in-the-loop to collect and verify human-written perturbations to benchmark toxic text detection, called \textsc{NoisyHate}, that consists of clean data and its corresponding perturbed version with online human-written text perturbations. The dataset allows researchers to evaluate the effectiveness of their proposed toxic content detection models in the face of real-world human-written perturbations instead of traditional machine-generated perturbations via existing adversarial text attacks. Furthermore, the diverse types of perturbations in the dataset pose a challenge for normalization algorithms, calling for better defenses to protect against potential adversarial behaviors online.

% \newpag

\section{Acknowledgements}

This work was in part supported by NSF awards \#1934782 and
\#2131144.

\bibliography{aaai25}

\newpage
\section{Ethical Checklist}
\begin{enumerate}

\item For most authors...
\begin{enumerate}
    \item  Would answering this research question advance science without violating social contracts, such as violating privacy norms, perpetuating unfair profiling, exacerbating the socio-economic divide, or implying disrespect to societies or cultures?
    {Yes}
  \item Do your main claims in the abstract and introduction accurately reflect the paper's contributions and scope?
    {Yes}
   \item Do you clarify how the proposed methodological approach is appropriate for the claims made?
    {Yes}
   \item Do you clarify what are possible artifacts in the data used, given population-specific distributions?
    {Yes}
  \item Did you describe the limitations of your work?
    {Yes}
  \item Did you discuss any potential negative societal impacts of your work?
    {Yes}
      \item Did you discuss any potential misuse of your work?
    {Yes}
    \item Did you describe steps taken to prevent or mitigate potential negative outcomes of the research, such as data and model documentation, data anonymization, responsible release, access control, and the reproducibility of findings?
    {Yes}
  \item Have you read the ethics review guidelines and ensured that your paper conforms to them?
    {Yes}
\end{enumerate}

\item Additionally, if your study involves hypotheses testing...
\begin{enumerate}
  \item Did you clearly state the assumptions underlying all theoretical results?
    {NA}
  \item Have you provided justifications for all theoretical results?
    {NA}
  \item Did you discuss competing hypotheses or theories that might challenge or complement your theoretical results?
    {NA}
  \item Have you considered alternative mechanisms or explanations that might account for the same outcomes observed in your study?
    {NA}
  \item Did you address potential biases or limitations in your theoretical framework?
    {NA}
  \item Have you related your theoretical results to the existing literature in social science?
    {NA}
  \item Did you discuss the implications of your theoretical results for policy, practice, or further research in the social science domain?
    {NA}
\end{enumerate}

\item Additionally, if you are including theoretical proofs...
\begin{enumerate}
  \item Did you state the full set of assumptions of all theoretical results?
    {NA}
	\item Did you include complete proofs of all theoretical results?
    {NA}
\end{enumerate}

\item Additionally, if you ran machine learning experiments...
\begin{enumerate}
  \item Did you include the code, data, and instructions needed to reproduce the main experimental results (either in the supplemental material or as a URL)?
    {Yes}
  \item Did you specify all the training details (e.g., data splits, hyperparameters, how they were chosen)?
    {NA}
     \item Did you report error bars (e.g., with respect to the random seed after running experiments multiple times)?
    {Yes}
	\item Did you include the total amount of compute and the type of resources used (e.g., type of GPUs, internal cluster, or cloud provider)?
    {NA}
     \item Do you justify how the proposed evaluation is sufficient and appropriate to the claims made?
    {Yes}
     \item Do you discuss what is ``the cost`` of misclassification and fault (in)tolerance?
    {NA}

\end{enumerate}

\item Additionally, if you are using existing assets (e.g., code, data, models) or curating/releasing new assets, \textbf{without compromising anonymity}...
\begin{enumerate}
  \item If your work uses existing assets, did you cite the creators?
    {Yes}
  \item Did you mention the license of the assets?
    {Yes}
  \item Did you include any new assets in the supplemental material or as a URL?
    {Yes}
  \item Did you discuss whether and how consent was obtained from people whose data you're using/curating?
    {Yes}
  \item Did you discuss whether the data you are using/curating contains personally identifiable information or offensive content?
    {Yes}
\item If you are curating or releasing new datasets, did you discuss how you intend to make your datasets FAIR (see \citet{fair})?
{Yes}
\item If you are curating or releasing new datasets, did you create a Datasheet for the Dataset (see \citet{gebru2021datasheets})?
{No}
\end{enumerate}

\item Additionally, if you used crowdsourcing or conducted research with human subjects, \textbf{without compromising anonymity}...
\begin{enumerate}
  \item Did you include the full text of instructions given to participants and screenshots?
    {Yes}
  \item Did you describe any potential participant risks, with mentions of Institutional Review Board (IRB) approvals?
    {Yes}
  \item Did you include the estimated hourly wage paid to participants and the total amount spent on participant compensation?
    {Yes}
   \item Did you discuss how data is stored, shared, and deidentified?
   {Yes}
\end{enumerate}

\end{enumerate}

\renewcommand{\tabcolsep}{2pt}
\begin{table*}[h]
    \centering
    \small
    % \scriptsize
    % \resizebox{\textwidth}{!}{
    \begin{tabular}{ll}
        \toprule
        \textbf{Section} & \textbf{Instruction} \\
        \cmidrule(lr){1-2}
        \multirow{6}{*}{\textbf{Definition}} & \textbf{What is a human written text perturbation?}\\
          &  A human written text perturbation is an alternative of a standard English \\
          & word that is usually invented by netizens. The purpose of a perturbation can be:\\
          &  1. avoiding platforms’ filtering: ass ${\rightarrow}$ a55; Russian ${\rightarrow}$ Russ!an\\
          &  2. visual attraction: bro ${\rightarrow}$ brooooooo; porn ${\rightarrow}$ pornnnnnn\\
          &  3. memes: republican + lie ${\rightarrow}$ repubLIEcan; democrat ${\rightarrow}$ democRAT\\
        \cmidrule(lr){1-2}
        \multirow{14}{*}{\textbf{Examples}} & \textbf{Perturbations that are \textcolor{green}{LIKELY} to be written by human:}\\
                               & Example 1:\\
                               & Clean Sentence: keeping \ul{russian} invaders warm in these frigid temperatures\\
                               & Perturbed Sentence: keeping \ul{russ!an} invaders warm in these frigid temperatures\\
                               & Example 2:\\
                               & Clean Sentence: our \ul{country} is a cesspool\\
                               & Perturbed Sentence: our \ul{countrrry} is a cesspool\\
                               & \textbf{Perturbations that are \textcolor{red}{UNLIKELY} to be written by human:}\\
                               & Example 1:\\
                               & Clean Sentence: \ul{damn} you beat me to it\\
                               & Perturbed Sentence: \ul{dawn} you beat me to it\\
                               & Example 2:\\
                               & Clean Sentence: they were pretty \ul{cool} to me when i was there\\
                               & Perturbed Sentence: they were pretty \ul{coll} to me when i was there\\
        \bottomrule
    \end{tabular}
    % }
    \caption{Guideline for MTurk Workers}
    \label{tab:instruction}
\end{table*}

% \newpage
% \newpage
% \pagebreak
\section{Appendix}

\end{document}